\documentclass[letterpaper, 10 pt, conference]{ieeeconf}  

\usepackage[T1]{fontenc}
\usepackage{graphicx}
\graphicspath{{Figs/}} 
\usepackage{amsmath,amsfonts}
\usepackage{algorithm}
\usepackage{algpseudocode}
\usepackage{booktabs}
\usepackage{pifont}  
\usepackage{tabularx}  
\newcolumntype{Y}{>{\centering\arraybackslash}X}

\makeatletter
\renewcommand{\fnum@figure}{Figure~\thefigure}
\makeatother

\title{\LARGE \bf 
Efficient Multi-Modal Planning with Reward-Guided Preference Optimization for Autonomous Driving}

\author{Chenglin Chen, Lujia Wang, Xinhu Zheng, Jun Ma, Haoang Li$^{*}$
\thanks{$^{1}$The Hong Kong University of Science and Technology (Guangzhou)}
}

\begin{document}

\maketitle
\thispagestyle{empty}
\pagestyle{empty}


\begin{abstract}
Safe and efficient trajectory planning is essential in autonomous driving. However, existing end-to-end approaches often fall short in both computational efficiency and safety guarantees. Methods based on imitation learning suffer from causal confusion, while rule-based scoring approaches often incur heavy computational overhead and suffer from objective misalignment. Additionally, preference-based methods rely on strict pairwise annotations, limiting data utilization.
To overcome these limitations, we propose EMPlan, an efficient multi-modal trajectory planning method powered by reward-guided fine-tuning. 
We design a hybrid architecture that combines sparse anchors with an offset refinement module for efficient multi-modal trajectory prediction. Sparse anchors provide coarse trajectory candidates with low latency, which are subsequently refined by the offset module for higher prediction accuracy. To enhance safety without incurring additional inference costs, we adopt a two-stage training paradigm consisting of pretraining and reward-guided fine-tuning. During fine-tuning, we leverage rule-based reward signals  and unpaired preference supervision to refine the pretrained policy toward safer trajectory selection. 
We evaluate EMPlan on the non-reactive NAVSIM benchmark, where it strikes a favorable balance between planning accuracy and efficiency, demonstrating superior performance under real-time constraints.
\end{abstract}


\section{Introduction}
Ensuring safety, comfort, and compliance with traffic rules are the fundamental objectives for an autonomous driving (AD) system. Recently, the end-to-end (E2E) AD has emerged as a promising solution to achieve these goals~\cite{hu2022st,chitta2022transfuser,hu2023planning,chen2024vadv2}. This paradigm allows the entire system to be optimized jointly, thereby eliminating the information loss inherent in traditional modular designs. 
Nevertheless, the training objective of most existing E2E systems remains rooted in imitation learning (IL), where policies are trained in an open-loop manner by simply aligning predicted trajectories with expert data~\cite{chen2022learning,bansal2018chauffeurnet}. IL-based methods often fail in long-tail situations due to causal confusion~\cite{geirhos2020shortcut} and distribution shifts~\cite{lu2023imitation,gao2025rad,ross2011reduction} in the training data. Furthermore, the objective of IL primarily aims to mimic expert data, often neglecting critical driving requirements such as safety, comfort, and adherence to traffic rules.
\begin{figure}[!t]
\centering
\includegraphics[width=0.45\textwidth]{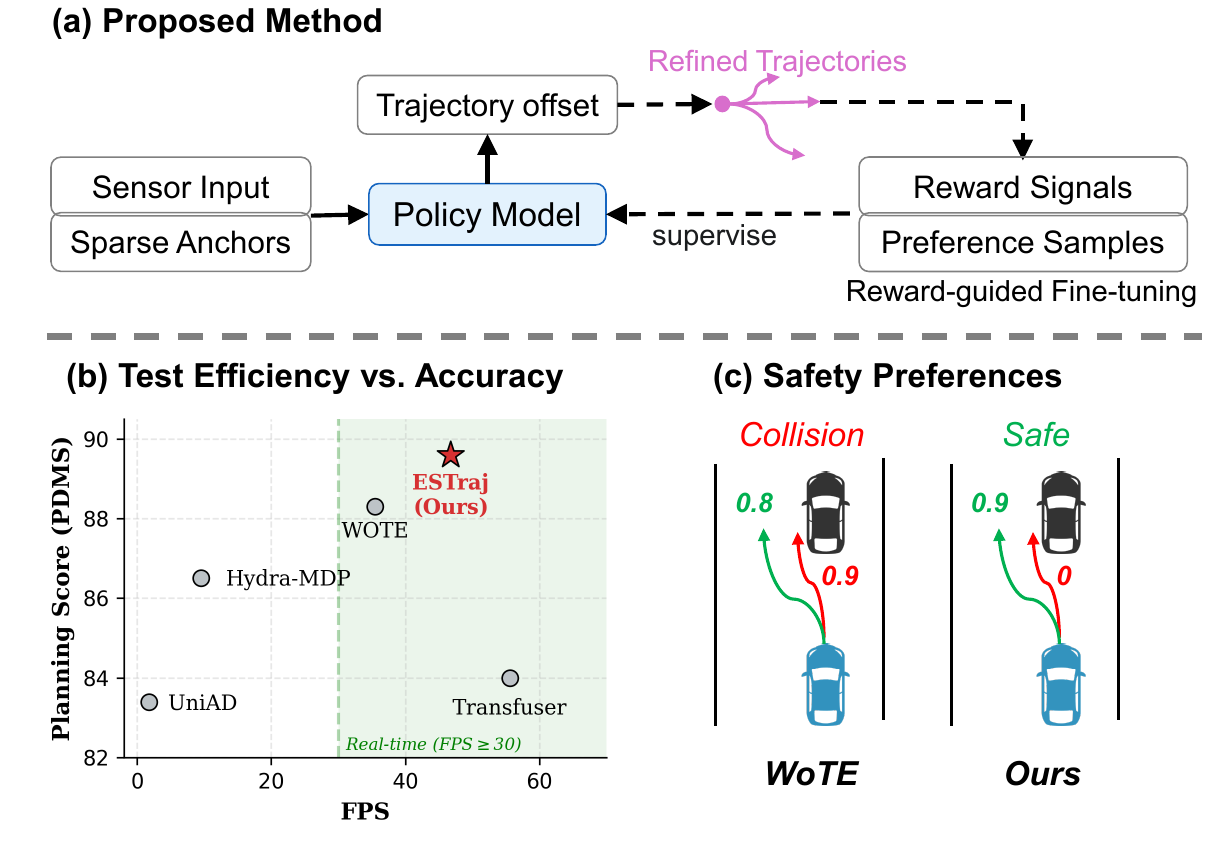}
\caption{Overview of our EMPlan, built upon a hybrid architecture and powered by reward-guided fine-tuning. (a) The hybrid architecture combines sparse anchors with an offset module for efficient multi-modal trajectory prediction. Reward-guided fine-tuning further aligns trajectory selection with safety preferences. (b) Our model achieves better performance in efficiency and accuracy compared with representative methods~\cite{chitta2022transfuser,hu2023planning,li2024hydra,li2025end}. (c) Compared to score-based methods (i.e., WoTE~\cite{li2025end}), our model assigns lower scores to human-like but unsafe trajectories (red) and higher scores to safe ones (green).}
\label{fig1}
\end{figure}

To alleviate these limitations, recent approaches attempt to refine IL-pretrained policies by incorporating auxiliary signals, such as rule-based scores~\cite{li2024hydra,li2025end} or preference-based labels~\cite{christiano2017deep,ouyang2022training,xiao2024comprehensive,rafailov2023direct,shang2025drivedpo}, to explicitly align the model with safe or logical driving behaviors. Although these approaches are generally more reliable than pure imitation, they still suffer from limitations in computational efficiency and safety guarantees. 
First, methods incorporating rule-based scores typically adopt anchor-based architecture. To ensure sufficient action space coverage, they require a dense set of anchors, resulting in prohibitive computational latency. Moreover, directly regressing rule-based scores often causes objective misalignment, as it encourages the model to fit absolute values rather than learning preference relationships necessary for reliable trajectory selection.
Second, preference-based methods attempt to better capture driving logic through trajectory-level comparisons. However, they generally rely on strict pairwise annotations (e.g., chosen vs. rejected) within each scenario. Such pairwise supervision is sparse and data-inefficient, as learning is restricted to limited trajectory pairs while other valid trajectory candidates are ignored. 

To overcome these limitations, we propose EMPlan (see Fig.~\ref{fig1}), a multi-modal trajectory planning method that unifies efficient trajectory generation with safety-aware trajectory selection. Our model is built upon a hybrid architecture and enhanced by reward-guided fine-tuning.
First, we design a hybrid architecture to improve efficiency. Unlike traditional methods that exhaustively enumerate thousands of anchors, we utilize sparse anchors to generate coarse trajectory candidates, which are then refined by a lightweight offset module. This hybrid design preserves action space coverage while significantly reducing computational overhead.
Second, to enhance safety without incurring additional inference costs, we employ a two-stage training paradigm consisting of pretraining and reward-guided fine-tuning. 
In the pretraining stage, we design a score-based policy learning objective that integrates trajectory imitation, rule-based supervision, and auxiliary BEV segmentation to better handle data imbalance and improve representation learning. 
In the fine-tuning stage, we optimize the pretrained policy using reward signals instead of rule-based scores. 
This aligns the training objective with the trajectory selection criterion at inference, reducing the mismatch between optimization and decision-making. 
However, reward supervision typically operates on individual trajectories and fails to explicitly model relative comparisons between candidates. We therefore incorporate preference-based signals to capture relative relationships within each scenario. To alleviate the reliance on strict pairwise annotations in preference learning, we utilize unpaired trajectory samples with preference annotations (e.g., desirable and undesirable samples). This approach improves the utilization of available trajectory data, aligning the policy with safe driving behaviors while avoiding the constraints of strict pairwise supervision.

In summary, our main contributions are as follows:
\begin{itemize}
\item We propose EMPlan, an efficient multi-modal trajectory planning method powered by reward-guided fine-tuning, which can achieve real-time inference and improve overall planning accuracy.
\item We design a hybrid architecture combining sparse anchors with an offset module to refine trajectories, overcoming discrete coverage gaps at low computational cost.  
\item   We introduce an unpaired, reward-guided fine-tuning strategy that adapts preference learning to anchor-based trajectory evaluation, enabling effective policy alignment with multi-objective driving criteria.
\end{itemize}

\section{Related Works}
\subsection{End-to-End Autonomous Driving}
Benefiting from large-scale driving datasets and advances in transformer architectures~\cite{vaswani2017attention}, E2E AD has progressed rapidly in recent years. Early works, such as Transfuser~\cite{chitta2022transfuser}, emphasize multi-modal sensor fusion to learn stronger scene representations. Moving towards planning-oriented design, UniAD~\cite{hu2023planning} unifies full-stack tasks within a single network to enhance planning performance. To further improve efficiency, VAD~\cite{jiang2023vad} explores compact vectorized scene representations. 
Despite architectural differences, these approaches typically regress a single-mode trajectory conditioned on the ego vehicle. To address the inherent multimodality of driving behaviors, VADv2~\cite{chen2024vadv2} introduces a predefined anchor vocabulary and learns to assign scores to candidate trajectories, enabling multi-mode prediction. Building upon this idea, HydraMDP~\cite{li2024hydra} distills rule-based knowledge from simulators into the model to enhance trajectory scoring. WoTE~\cite{li2025end} leverages a world model to generate future BEV representations as intermediate supervision, improving downstream trajectory prediction. Nevertheless, such score-based models are optimized by assigning supervision scores to each anchor in isolation, while ignoring the relative preference relations among different trajectory candidates. More recently, diffusion models have been explored for multi-modal trajectory prediction. Despite their strong generative capability, diffusion-based approaches~\cite{li2024drivingdiffusion,liao2025diffusiondrive} typically incur high memory overhead and inference latency, which hinder their deployment in real-time safety-critical AD systems. In contrast, we propose a lightweight hybrid architecture model for multimodal trajectory prediction, further enhanced by reward-guided fine-tuning to encourage safer planning behaviors.

\subsection{Policy Fine-Tuning in Autonomous Driving}
Recently, policy fine-tuning has emerged as a promising direction for improving E2E AD, mainly through closed-loop reinforcement learning (RL) and preference-based optimization. RL-based methods benefit from interactive environment signals but are still in an early stage for AD, as they rely heavily on domain-specific reward engineering and often exhibit unstable training. 
To bridge the simulation-to-reality gap, RAD~\cite{gao2025rad} introduces a 3D Gaussian Splatting-based RL framework for training in high-fidelity environments. In parallel, both AlphaDrive~\cite{jiang2025alphadrive} and TrajHF~\cite{li2025finetuning} adopt Group Relative Policy Optimization~\cite{shao2024deepseekmath} for policy optimization, where the former incorporates vision-language reasoning, and the latter aligns diffusion planners with human driving preferences. Furthermore, ReCogDrive~\cite{li2025recogdrive} leverages simulator-assisted RL to refine trajectory generation toward safer and more human-like behaviors. 
More recently, DriveDPO~\cite{shang2025drivedpo} fine-tunes anchor-based planners via DPO, but it relies on strict pairwise preference annotations, which may leave valuable unpaired trajectories underutilized. 
In contrast, our method adopts unpaired preference data, allowing each candidate trajectory to contribute independently to preference optimization. Our goal is to improve the alignment between predicted scores and actual driving quality without requiring explicit pair construction. This is particularly beneficial under a multi-candidate selection setting, where candidate trajectories can be independently evaluated rather than selectively paired.

\begin{figure*}[!t]
\centering
\includegraphics[width=0.85\textwidth]{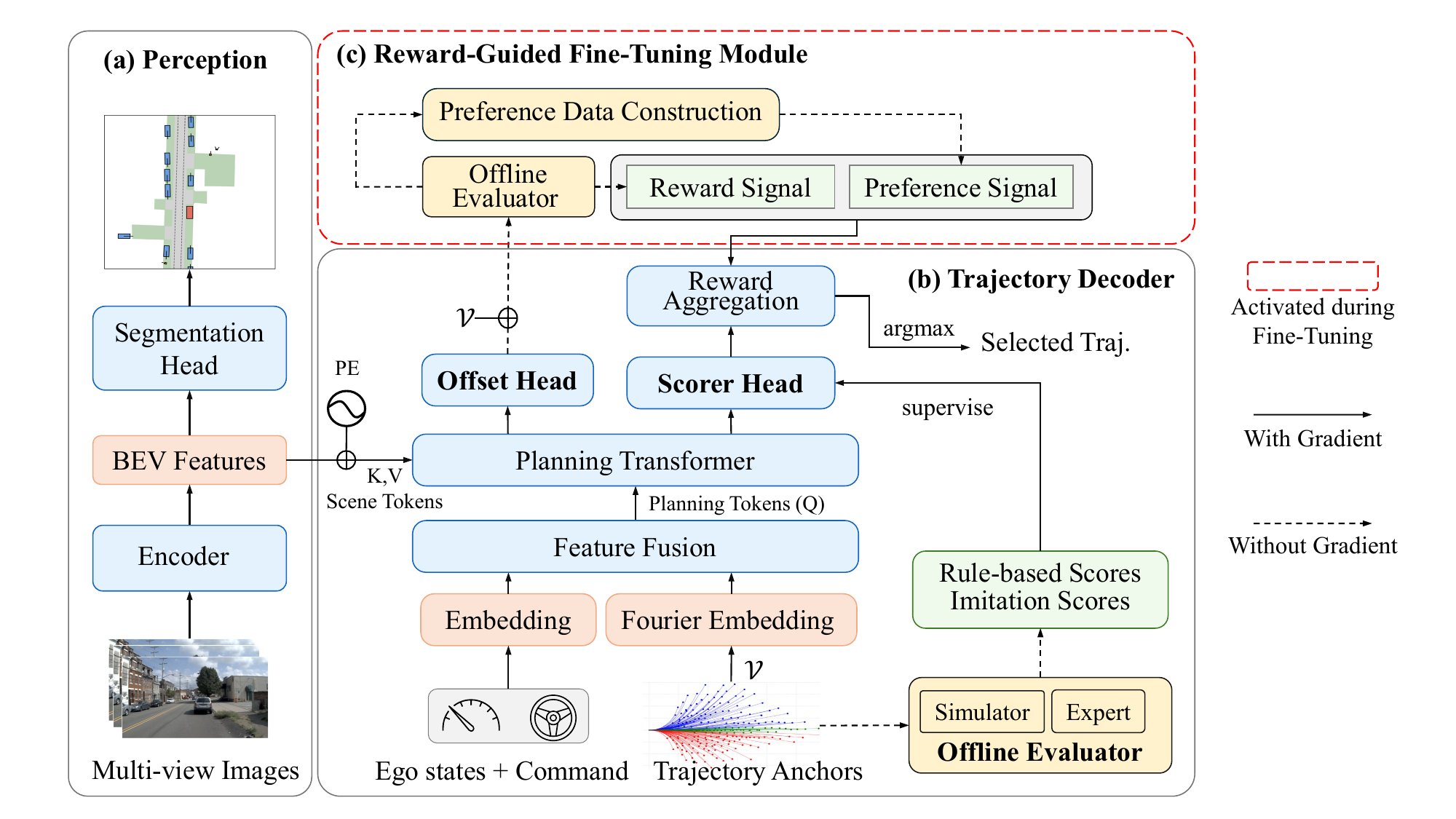}
\caption{Overview of the network architecture. Given multi-view images, the perception module extracts BEV representations, which are fed into a trajectory decoder to predict trajectory offsets and reward scores based on ego states and sparse anchors. The offset head refines anchor trajectories, and the trajectory with the highest aggregated reward is chosen as the final output. During training, the model is first pretrained with imitation and rule-based supervision and subsequently optimized via reward-guided fine-tuning, which integrates reward loss and preference loss to align trajectory selection with safety preferences.}
\label{overview}
\end{figure*}

\section{Method}
Given the sensor observation $\mathcal{O}$, the goal of the model is to forecast a $T$-step ego-vehicle’s future trajectory $\tau=\{(x_t,y_t,\theta_t)\}_{t=1}^T$, where each waypoint denotes the vehicle’s position and headings at time step $t$. 
To handle the multi-modal nature of real-world driving, we follow the paradigm in VADv2~\cite{chen2024vadv2} and treat motion planning as a discrete selection problem over a predefined vocabulary of trajectory anchors $\mathcal{V}=\{\tau_i\in \mathbb{R}^{T\times 3} \}_{i=1}^N$, where each anchor $\tau_i$ represents a $T$-step waypoint sequence. 
The policy assigns a score to each anchor, and the anchor with the highest score is selected as the predicted trajectory.

Standard supervised training optimizes these scores independently for each anchor, which ignores relative quality among candidates and requires a dense anchor set to cover the action space, leading to high computational cost. To address these limitations, we aim to 1) efficiently generate trajectories without exhaustive anchor enumeration and 2) leverage relative reward signals to refine the policy toward safer trajectory selection. The following sections detail the overall architecture and key optimization strategies.

\subsection{Hybrid Architecture of Our Policy Model}
As shown in Fig.~\ref{overview},  our method is structured into two networks: a Perception module for scene understanding and a Trajectory Decoder for multimodal trajectory generation. 
For the perception module, we extract features from multi-view RGB images and transform them into BEV representations at two distinct scales. One scale is utilized for BEV semantic segmentation with auxiliary supervision to enhance spatial awareness. The other is combined with learnable positional embeddings to form scene tokens $E_{\textnormal{env}}$, which provide rich, scene-level context for the subsequent trajectory decoder. 
The core of our architecture lies in the design of the trajectory decoder.
To balance computational efficiency and planning accuracy, we decompose trajectory generation into coarse discrete mode representation and continuous offset refinement. The sparse anchors represent distinct maneuver modes, while the offset head predicts a scene-conditioned residual for each anchor, reducing the quantization error imposed by a finite anchor vocabulary. In contrast, dense-anchor methods obtain finer discrete coverage by increasing the number of anchors, which enlarges the candidate vocabulary and raises the computation of trajectory decoding. Unlike diffusion planners, our refinement is completed in a single feed-forward pass without iterative denoising.

Specifically, we begin by constructing a sparse set of trajectory anchors $\mathcal{V}=\{\tau_i\in \mathbb{R}^{T\times 3} \}_{i=1}^N$ by applying k-means clustering to the NAVSIM training dataset, following prior works~\cite{chen2024vadv2,li2024hydra,li2025end}. 
These anchors are transformed into a high-dimensional feature space through Fourier positional encoding to obtain the anchor embeddings:
\begin{subequations}
\begin{equation}
\Gamma(\mathcal{V}) = 
\Big[ 
\mathcal{V}, \phi_0 (\mathcal{V}), \phi_1 (\mathcal{V}), \cdots, \phi_{L-1} (\mathcal{V}) 
\Big] 
\in \mathbb{R}^{N \times T \times (3+6L)}
\end{equation}
\begin{equation}
\phi_k (\mathcal{V}) = \big[ \sin(2^k \pi \mathcal{V}), \cos(2^k \pi \mathcal{V}) \big]
\end{equation}
\end{subequations}
In addition, the ego state and high-level driving command are also encoded and fused with these anchor embeddings to form the final planning tokens $E_{\textnormal{plan}}$ for trajectory decoding. 
The scene tokens $E_{\textnormal{env}}$ and planning tokens $E_{\textnormal{plan}}$  are then fed into a Planning Transformer to perform cross-attention, producing a set of scene-conditioned planning tokens:
\begin{equation}
\hat{E}_{\textnormal{plan}} = \mathrm{TransformerDecoder} \left(Q=E_{\textnormal{plan}}, K,V=E_{\textnormal{env}} \right)
\end{equation}
Finally, the context-aware planning tokens $\hat{E}_{\textnormal{plan}}$  are passed to two separate prediction heads: an offset head and a scorer head. 
Since the predefined trajectory anchors form a sparse and discrete motion prior, the ground-truth trajectories may not perfectly align with them. To address this, an offset head predicts per-anchor offsets to refine the coarse trajectories. The refined trajectories are given by:
\begin{equation}
 \hat{\mathcal{V}} = \mathcal{V} + \mathrm{OffsetHead} \left( \hat{E}_{\textnormal{plan}} \right)
\end{equation}
In parallel with trajectory refinement, the scorer head evaluates the quality of each refined trajectory. Specifically, it outputs an imitation-based score $S_{\textnormal{im}}$ to measure human-likeness and a set of rule-based scores $S_{\textnormal{sim}}$ to assess safety and comfort constraints. These per-anchor scores are further unified via a reward aggregation module into a single reward for each trajectory. 
During inference, the optimal trajectory is chosen by maximizing the predicted reward.
During training, the predicted scores together with the refined trajectories are supervised to optimize the policy in the first stage.


\subsection{Score-Based Policy Learning}
This section details the supervision and loss formulation for first-stage policy learning. Following established practices~\cite{li2024hydra,zhou2023query}, we employ a trajectory regression loss $\mathcal{L}_{\textnormal{traj}}$ with a winner-takes-all strategy and an imitation-based score loss $\mathcal{L}_{\textnormal{im}}$ to encourage human-like trajectory prediction.

\textbf{BEV Segmentation Loss.} 
We introduce an auxiliary BEV semantic segmentation task to enhance scene representation. The BEV map contains multiple categories (e.g., background, road, lane centerline, vehicles) with severe class imbalance, as background and road dominate the scene. Unlike prior work~\cite{chitta2022transfuser,li2024hydra} that adopts cross-entropy loss, we supervise the predicted map using focal loss $\mathcal{L}_{\textnormal{map}}$ to emphasize hard and minority categories.

\textbf{Rule-Based Score Loss. }To enforce safety and physical constraints beyond imitation, we supervise the predicted rule-based scores $S_{\textnormal{sim}}$ using precomputed outputs $S_{\textnormal{sim}}^*$ from NAVSIM simulator~\cite{dauner2024navsim}. The simulator provides five rule-based metrics: No At-Fault Collision (NC), Drivable Area Compliance (DAC), Time-to-Collision (TTC), Comfort (C), and Ego Progress (EP). The distribution of rule-based scores is heavily skewed, as the majority of anchors correspond to low-quality behaviors. To mitigate this imbalance, we adopt a Sigmoid Focal Loss (SFLoss) for supervision. Notably, for the EP score, we introduce a conditional progress mask. Since EP measures agent progress but is set to zero when either NC or DAC is zero, the mask ensures that EP loss is applied only for trajectories that are safe and legal. This prevents the model from being misled by unsafe behavior while still learning meaningful progress signals. As a result, the loss is defined as:
\begin{equation}
\mathcal{L}_{\textnormal{sim}} = 
\begin{cases}
{\mathrm{Mask}_{\textnormal{EP}} \cdot 
\textnormal{SFLoss}\left(S_{\textnormal{sim}}, S_{\textnormal{sim}}^{*}\right),}
&{\text{if}}\ \textnormal{sim} = \textnormal{EP} \\
\textnormal{SFLoss}\left(S_{\textnormal{sim}}, S_{\textnormal{sim}}^{*}\right),
&{\text{otherwise.}} 
\end{cases} \\
\end{equation}

\noindent where $\mathrm{Mask}_{\textnormal{EP}}=(S_{\textnormal{NC}} > 0.5) \wedge (S_{\textnormal{DAC}} > 0.5)$.
Finally, the total loss in the first stage is defined as:
\begin{equation}
\mathcal{L}_{total} =\lambda_{\textnormal{map}} \mathcal{L}_{\textnormal{map}} + \lambda_{\textnormal{traj}}\mathcal{L}_{\textnormal{traj}}+ \lambda_{\textnormal{im}}\mathcal{L}_{\textnormal{im}} + \lambda_{\textnormal{sim}}\mathcal{L}_{\textnormal{sim}}
\end{equation}

\noindent where $\lambda_{\textnormal{map}}, \lambda_{\textnormal{traj}}, \lambda_{\textnormal{im}}, \lambda_{\textnormal{sim}}$ balance the contributions of the respective loss terms.

\subsection{Reward-Guided Fine-Tuning}
\begin{figure}[!t]
\centering
\includegraphics[width=0.47\textwidth]{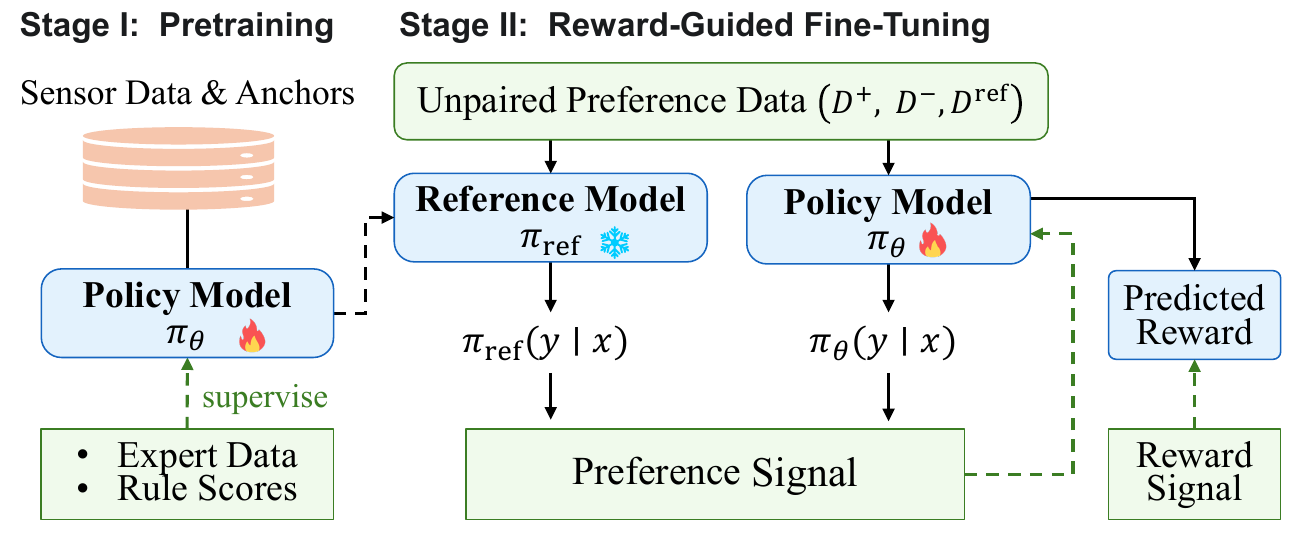}
\caption{Illustration of our two-stage training pipeline. In Stage I, $\pi_\theta$ is pretrained using expert demonstrations and rule-based scores to learn human-like driving priors. In Stage II, rule-based scores are combined to compute a trajectory-level reward loss across all candidates. In addition, unpaired preference data are used to compute preference loss (see Eq. (\ref{lkto})), enabling the model to learn safer and more reliable trajectory selection.}
\label{flow}
\end{figure}

The first-stage policy learning yields a pretrained reference policy $\pi_\textnormal{ref}$, which predicts per-anchor scores and refined trajectories. However, regressed absolute scores can be noisy and fail to explicitly model relative preferences among candidate trajectories. In reality, driving decisions rely more on relative preferences than absolute scoring. To bridge this gap, we introduce a reward-guided fine-tuning strategy (see Fig.~\ref{flow}) that combines two supervision signals: i) reward supervision over all candidate trajectories and ii) preference-based optimization by unpaired trajectory samples with preference annotations. Preference-based optimization captures trajectory preferences but is prone to training instability and distribution drift when applied alone. Inspired by~\cite{lu2023imitation}, we introduce reward supervision as an auxiliary loss to mitigate this issue, which also helps to keep the training objective consistent with trajectory selection. In particular, we break down this process into two key components: reward design and preference-based optimization in Stage II.

\textbf{Reward Design.} During reward-guided fine-tuning, we employ the reward aggregation module to unify the per-anchor scores predicted by the policy model into a single scalar reward for each trajectory.  For each candidate trajectory \(y\) under scene \(x\), the reward \(R(x,y)\) is computed as a weighted combination of the imitation score $S_{\textnormal{im}}$ and rule-based scores $ S_{\textnormal{sim}} \in \{{S_{\textnormal{NC}}, S_{\textnormal{DAC}}, S_{\textnormal{EP}}, S_{\textnormal{TTC}}, S_{\textnormal{C}}}\}$ in log space:
\begin{align}
R(x,y) = &\omega_1 \log S_{\textnormal{im}} +
\omega_2 \log S_{\textnormal{NC}} +\omega_3 \log S_{\textnormal{DAC}} \nonumber \\
&+ \omega_4 \log \frac{5S_{\textnormal{EP}} +5S_{\textnormal{TTC}}+2S_{\textnormal{C}} }{12} 
\end{align}
\noindent where $\{ \omega_i\}_{i=1}^4$ are the hyperparameters that balance driving style and safety. 
We use log-space aggregation to reflect the conjunctive nature of the safety-related criteria, so that high EP, TTC, or C scores cannot compensate for near-zero NC or DAC scores. Unlike a conventional weighted sum, which allows compensation across all components, the weighted sum of logarithms is equivalent to the logarithm of a weighted geometric product. Consequently, when any individually logged factor approaches zero, the aggregated score decreases sharply, strongly penalizing the corresponding trajectory.
These scores serve as unified reward signals for both direct supervised training and subsequent preference-based optimization.

\begin{figure}[!t]
\centering
\includegraphics[width=0.47\textwidth]{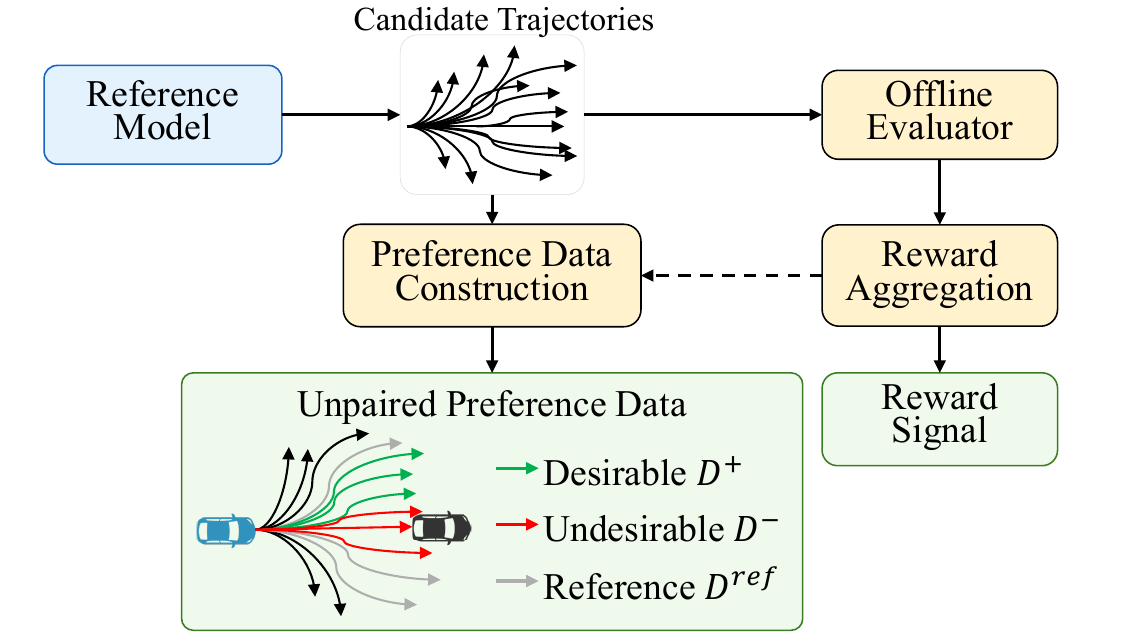}
\caption{Illustration of our construction of reward signal and preference data. The policy model pretrained in Stage I first generates candidate trajectories, which are then evaluated by an offline evaluator to obtain reward scores. The resulting rewards are used both for dense reward supervision and for constructing unpaired preference data by selecting $\mathcal{D}^{+}$, $\mathcal{D}^{-}$, $\mathcal{D}^\textnormal{ref}$.}
\label{ft_signal}
\end{figure}

\textbf{Preference-Based Optimization.} To further align the policy with robust driving behaviors, we introduce Kahneman–Tversky Optimization (KTO) ~\cite{ethayarajh2024model}, which encourages desirable outputs while penalizing undesirable ones using unpaired samples rather than strict pairwise data.
To enable preference-based fine-tuning, our first step is to construct preference samples.  As illustrated in Fig.~\ref{ft_signal}, we utilize the pretrained model $\pi_\textnormal{ref}$ to generate multiple trajectories and feed them into an offline evaluator to obtain trajectory-level imitation scores, rule-based scores, and aggregated rewards.
We restrict preference partitioning to the top 25\% of candidates ranked by L2 similarity to the expert demonstration, since the remaining trajectories deviate substantially from expert behavior and are therefore excluded from fine-grained preference construction.
These human-like trajectories are then partitioned into three disjoint sets according to two empirically defined reward thresholds, $\tau_{\textnormal{high}}=-0.6$ and $\tau_\textnormal{low}=-1.0$.
Trajectories with rewards above the threshold $\tau_\textnormal{high}$ form the desirable set $\mathcal{D}^{+}$, corresponding to strictly safe behaviors. Those with rewards below the threshold  $\tau_\textnormal{low}$ form the undesirable set $\mathcal{D}^{-}$, corresponding to safety-violating behaviors. The remaining trajectories constitute the reference set $\mathcal{D}^\textnormal{ref}$, which contains safe yet suboptimal behaviors and is used to estimate the reference point as the average performance of the reference model. This partition enables the model to capture finer-grained preferences among trajectories with similar quality, providing more informative supervision for trajectory ranking than rule-based scores.

Second, we define a relative reward and formulate the corresponding loss accordingly. We use the relative reward $r_{\theta}(x,y)$ to measure the preference of the current policy for output $y$ relative to the reference policy under scene $x$. We denote $\pi_\theta(y\mid x)$ as the probability assigned to the trajectory $y$ by the current policy, obtained by normalizing the aggregated rewards $R(x,y)$ over the set of trajectory anchors $\mathcal{V}$ using a softmax operation:
\begin{equation}
\pi_{\theta}(y\mid x)
=
\frac{\exp\left(R(x,y')\right)}
{\sum_{y'\in\mathcal{V}}
\exp\left(R(x,y')\right)}
\end{equation}
Similarly, $\pi_{\textnormal{ref}}(y\mid x)$ denotes the corresponding probability assigned by the frozen reference policy. Thus, the relative reward is given by:
\begin{equation}
r_\theta(x,y) = \log {\pi_\theta(y\mid x)} - \log {\pi_{\textnormal{ref}}(y \mid x)}
\end{equation}
A larger $r_{\theta}(x,y)$ indicates that $\pi_\theta$ assigns a higher preference to $y$ compared to $\pi_\textnormal{ref}$, while a smaller value suggests the opposite.

We further estimate reference points to stabilize training. KTO was originally proposed for text generation, where the output space is continuous and many desirable samples are available. In contrast, anchor-based trajectory prediction operates over a discrete candidate set, in which a large portion of sampled trajectories may be infeasible or unsafe. Consequently, computing the reference point over all predicted trajectories can be biased, as low-quality samples may dominate the expectation. To mitigate this, we estimate the reference point $z_0$ as the average relative reward over the reference set:
\begin{equation}
\label{z0}
z_0 = \mathbb{E}_{y'\sim \mathcal{D}^\textnormal{ref}}\big[ r_\theta(x, y') \big]
\end{equation}
Finally, we formulate the  preference loss objective:

\begin{equation}
\label{lkto}
\begin{aligned}
\mathcal{L}_{\textnormal{KTO}} &=\mathbb{E}_{(x,y)}
\left[w(y)\left(1-v(x,y)\right)\right], \\
v(x,y) &= \begin{cases}
\sigma\left(\beta\left( r_\theta(x, y)-z_0 \right)\right),
 \text{if } y \in \mathcal{D}^{+} \\
\sigma\left(\beta\left(z_0 - r_\theta(x, y) \right)\right),
 \text{if } y\in \mathcal{D}^{-}
\end{cases}
\end{aligned}
\end{equation}

\noindent where $w(y)$ denotes the sample weight and is set to $\lambda_D$ or $\lambda_U$ for desirable or undesirable samples, respectively.  $\sigma(\cdot)$ denotes the sigmoid function, and $\beta>0$ controls the sensitivity of the objective to deviations of the relative reward from the reference point. This objective encourages the policy to increase the relative likelihood of desirable trajectories while suppressing undesirable ones with respect to the reference distribution. 

\begin{table*}[!t]
    \caption{Performance comparison with representative methods on the navtest split. 
    C: camera. L: LiDAR.}
    \label{tab_c1}
    \centering
    \begin{tabular}{cccc|ccccc||cc}
    \toprule
        Method &  Input&  Supervision
&Anchor 
&  NC$\uparrow$&  DAC$\uparrow$&  EP$\uparrow$&  TTC$\uparrow$&  C$\uparrow$& PDMS$\uparrow$ (Overall) \\
    \midrule
Human &  -&  -&- &  100.0&  100.0&  87.5& 100.0&  99.9& 94.8 \\
PDM-closed &  Perception GT&  - &-  &  94.6&  99.8&  89.9&  86.9&  99.9& 89.1  \\
\midrule
Transfuser~\cite{chitta2022transfuser}
&  C \& L&  Human &- &  97.7&  92.8&  79.2&  92.8&  \textbf{100.0}& 84.0 \\
UniAD~\cite{hu2023planning}
&  C&  Human &- &  97.8&  91.9&  78.8&  92.9&  \textbf{100.0}& 83.4 \\
VADv2~\cite{chen2024vadv2}
& C& Human &- & 97.9& 91.7& 77.6& 92.9& \textbf{100.0}& 83.0\\
PARA-Drive~\cite{weng2024drive}& C& Human &- & 97.9& 92.4& 79.3& 93.0& 99.8& 84.0\\
LAW~\cite{li2024enhancing} &  C&  Human &- &  96.4&  95.4&  81.7&  88.7&  99.9& 84.6\\
DRAMA~\cite{yuan2024drama}& C \& L&  Human &- & 98.0& 93.1& 80.1& 94.8& \textbf{100.0}& 86.5\\
 DiffusionDrive~\cite{liao2025diffusiondrive}& C \& L& Human & -& 98.2& 96.2& \textbf{94.7}& 82.2& \textbf{100.0}& 88.1\\
\midrule
Hydra-MDP~\cite{li2024hydra}
&  C \& L&  Human \& Rule &8192  &  98.3&  96.0&  78.7&  94.6&  \textbf{100.0}& 86.5 \\
WoTE~\cite{li2025end} & C \& L	& Human \& Rule
&256  
&\textbf{98.5}	&96.8	&81.9	&\textbf{94.9}	&99.9	&88.3\\
 EMPlan (Ours) & C& Human \& Rule&256 & 98.4& \textbf{97.5}& 84.2& \textbf{94.9}
&\textbf{100.0}&
\textbf{89.6}\\
\bottomrule
    \end{tabular}
\end{table*}

\begin{figure*}[!t]
\centering
\includegraphics[width=0.85\textwidth]{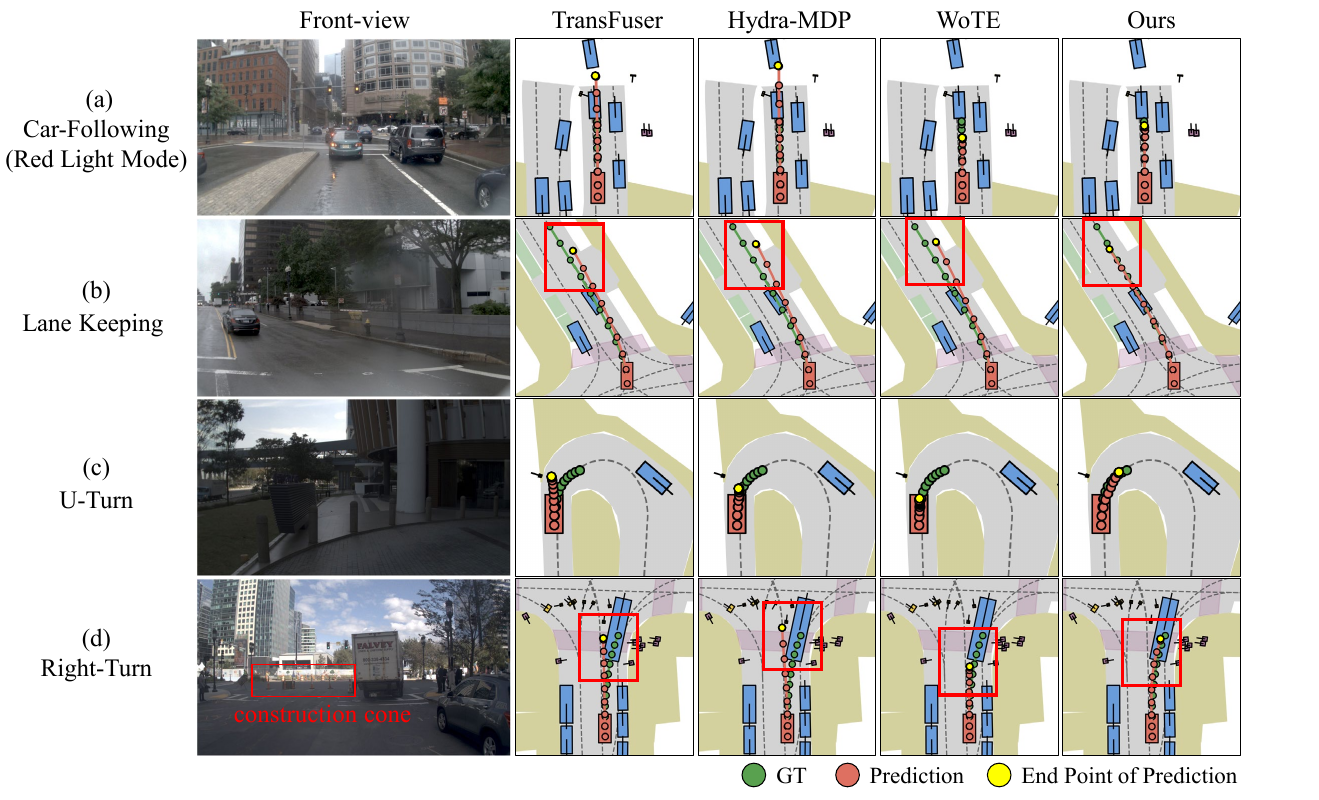}
\caption{Qualitative comparison of predicted trajectories across challenging scenarios. We present qualitative comparisons with TransFuser~\cite{chitta2022transfuser}, Hydra-MDP~\cite{li2024hydra}, and the SOTA WoTE~\cite{li2025end}. In (a), both TransFuser and Hydra-MDP fail to respond properly to the red traffic signal, causing rear-end collisions with the leading vehicle.
In addition, all baselines exhibit lane deviations in (b). 
WoTE tends to be overly conservative, resulting in stagnation behaviors in (c). In (d), TransFuser and Hydra-MDP proceed straight toward the construction obstruction, while WoTE avoids collision but fails to commit to the required right turn. In contrast, our EMPlan successfully executes all maneuvers.}
\label{v1}
\end{figure*}

\section{Experiment}
\subsection{Experimental Setup}\label{exp}
\textbf{Dataset}. We conduct experiments on NAVSIM~\cite{dauner2024navsim}, an E2E driving benchmark built from the OpenScene~\cite{peng2023openscene} redistributed version of nuPlan~\cite{caesar2021nuplan}. NAVSIM provides 2 Hz sensor inputs, along with semantic BEV maps and 3D bounding box annotations of objects. NAVSIM provides 1192 scenes in the navtrain split for training and validation, and 136 scenes in the navtest split for testing.

\textbf{Metrics}. Following prior work~\cite{li2024hydra,li2025end,liao2025diffusiondrive}, we use the Predictive Driving Model Score (PDMS) as the primary evaluation metric under a non-reactive simulation setting. PDMS is computed from five sub-metrics: NC, DAC, TTC, C, and EP. The final score is defined as
$\mathrm{PDMS} = \mathrm{NC}  \times \mathrm{DAC}  \times (5\mathrm{EP} + 5\mathrm{TTC} + 2\mathrm{C}) / 12$.

\textbf{Implementation Details.}
We use concatenated $1024 \times 256$ multi-view RGB images for perception input, following the commonly adopted setting in Transfuser~\cite{chitta2022transfuser}. We use V2-99~\cite{lee2019energy} as the backbone to extract image features. We set $N=256$ predefined trajectory anchors. The reward weights $w_2=w_3=0.5$, $w_4=1.0$ are set informed by the relative importance of components in the PDMS metric, and a smaller weight $w_1=0.1$ is assigned to the imitation term to prioritize safety over style imitation.
Training is performed on a platform with two NVIDIA RTX A6000 GPUs. For the first-stage training, we set $\lambda_{\textnormal{map}}=10.0$, $\lambda_{\textnormal{traj}}=1.0$, $\lambda_{\textnormal{im}}=1.0$, and $\lambda_{\textnormal{sim}} = \{\textnormal{NC}:5.0, \textnormal{DAC}:6.0, \textnormal{EP}:3.0, \textnormal{TTC}:3.0, \textnormal{C}:1.0\}$ to balance the loss term. We pretrain the model for 35 epochs with AdamW at a learning rate of $1\times 10^{-4}$ and a batch size of 16 per GPU. In the fine-tuning stage, we further optimize the policy for 10 epochs with a learning rate of $5\times 10^{-5}$ and a batch size of 16 per GPU in an iterative manner. During inference, the model predicts 8-waypoint trajectories over a 4-second horizon.

\subsection{Comparisons with State-of-the-art Methods}
\begin{table}[!b]
    \caption{Efficiency comparison between anchor-based methods. All models were tested on an NVIDIA RTX A6000.}
    \label{tab_efficiency}
    \centering
    \begin{tabular}{ccc}
    \toprule
    Method & PDMS$\uparrow$ & Latency$\downarrow$ \\
    \midrule
    Hydra-MDP~\cite{li2024hydra} & 86.5 & 104.7 ms \\
    WoTE~\cite{li2025end} & 88.2 & 28.3 ms \\
    EMPlan (Ours) & \textbf{89.6} & \textbf{21.4} ms \\
    \bottomrule
    \end{tabular}
\end{table}

\begin{table*}[!t]
  \centering
  \begin{minipage}[t]{0.25\textwidth}
    \caption{Ablation study for offset refinement and SFLoss.} 
    \label{a_design}
    \begin{tabularx}{\linewidth}{@{}YYY@{}} 
      \toprule
      offset & SFLoss & PDMS$\uparrow$ \\
      \midrule
      \ding{55} & \ding{55} & 81.6 \\
      \ding{51} & \ding{55} & 87.8 \\
      \ding{51} & \ding{51} & \textbf{88.3} \\
      \bottomrule
    \end{tabularx}
  \end{minipage}%
 \hfill
  \begin{minipage}[t]{0.32\textwidth}
    \caption{Analysis on number of predefined anchors $N$.} 
    \label{a_anchor}
    \begin{tabularx}{\linewidth}{@{}YYYY@{}}  
      \toprule
      $N$ & PDMS$\uparrow$&Latency$\downarrow$ & Param. \\
      \midrule
    128& 88.2 &\textbf{17.5} ms&  \\
    256&  88.3 & 21.4 ms&  113 M \\
    512&  \textbf{88.4}&24.8 ms  &  \\
      \bottomrule
    \end{tabularx}
  \end{minipage}%
\hfill
  \begin{minipage}[t]{0.3\textwidth}
     \caption{Ablation study of KTO with auxiliary supervision (Sup.).}
    \label{a_ft}
    \begin{tabularx}{\linewidth}{@{}YY@{}}
      \toprule
      Method & PDMS$\uparrow$ \\
      \midrule
      w/o Fine-tuning & 88.3 \\
      KTO + Score Sup. & 89.0 \\
      KTO + Reward Sup. & \textbf{89.6} \\
      \bottomrule
    \end{tabularx}
  \end{minipage}%
\end{table*}

To evaluate the effectiveness of EMPlan, we perform experiments on the NAVSIM navtest split against open-sourced representative planning baselines. These include the imitation-based Transfuser~\cite{chitta2022transfuser}, the diffusion-based DiffusionDrive~\cite{liao2025diffusiondrive}, and the scoring-based Hydra-MDP~\cite{li2024hydra} and WoTE~\cite{li2025end}, where WoTE represents the previous state-of-the-art (SOTA).
As shown in Table~\ref{tab_c1}, our model achieves the best PDMS of 89.6, indicating superior overall driving performance. 
Among the baselines, Transfuser (84.0) suffers from causal confusion and often produces unsafe behaviors in complex interactions, including collisions and drivable-area violations, as shown in Figs.~\ref{v1}(a, b). DiffusionDrive (88.1) achieves the highest EP but underperforms our method on critical safety indicators, particularly NC and DAC. This highlights that PDMS provides a more reliable evaluation of driving performance, whereas a sub-metric alone can be misleading.
Scoring-based methods such as Hydra-MDP (86.5) and WoTE (88.3) reduce collision rates but still struggle with consistent lane-keeping (see Fig.~\ref{v1}(b)), making them highly prone to off-road incidents. More importantly, their safety gains largely stem from overly conservative behavior, resulting in unnecessary braking and stagnated behavior as reflected by lower EP. As illustrated in the U-turn scenario in Fig.~\ref{v1}(c), both Hydra-MDP and WoTE fail to complete the maneuver and remain stalled, whereas our EMPlan successfully executes the turn. 
As further illustrated in the right-turn scenario with road construction in Fig.~\ref{v1}(d), Transfuser and Hydra-MDP fail to recognize the construction cones and proceed straight, resulting in collisions. Although WoTE avoids an immediate collision, continuing straight remains potentially unsafe given the obstruction ahead. In contrast, our EMPlan executes a smooth and decisive right turn, demonstrating its ability to suppress trajectories that appear human-like but carry latent safety risks.

Table~\ref{tab_efficiency} compares our EMPlan with publicly available representative anchor-based methods with respect to inference efficiency.
Notably, our method achieves the lowest latency (21.4 ms) while maintaining the highest PDMS score (89.6). This efficiency gain is primarily attributed to our hybrid architecture, which bypasses the heavy computational burden from dense anchors. In addition, EMPlan relies solely on camera inputs and operates at real-time inference speed, further enhancing its suitability for real-world deployment.

\subsection{Ablation Study}
Table~\ref{a_design} presents the ablation results of offset-based trajectory refinement and SFLoss. Adding offset results in a notable performance gain, improving PDMS from 81.6 to 87.8. This is primarily because the base model relies on sparse predefined anchors that cannot fully cover diverse driving behaviors, while offset refinement enables fine-grained trajectory adjustments beyond discrete anchor limitations. Incorporating SFLoss further increases PDMS to 88.3 by emphasizing hard samples during training.

Table~\ref{a_anchor} analyzes the effect of the number of predefined anchors $N$. As Nincreases from 128 to 512, PDMS improves slightly from 88.2 to 88.4, indicating that a denser anchor set provides better coverage of diverse driving behaviors. However, increasing $N$ also leads to noticeably higher inference latency (from 17.5 ms to 24.8 ms), which may hinder efficiency in closed-loop evaluation. Considering both accuracy and efficiency, we adopt $N=256$, as it provides a favorable trade-off between accuracy and efficiency.

\begin{figure}[!t]
\centering
\includegraphics[width=0.45\textwidth]{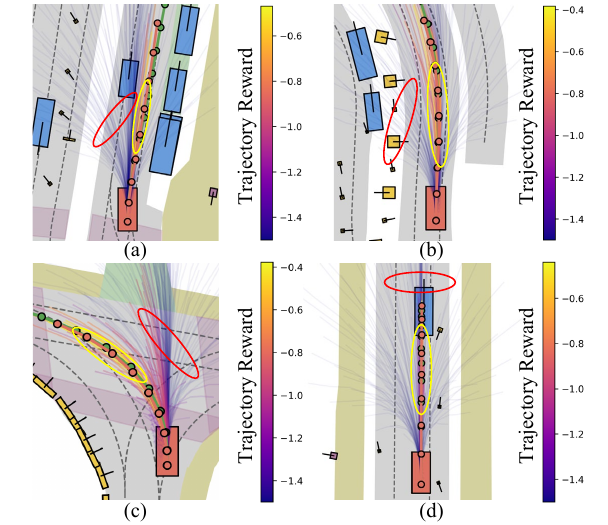}
\caption{Visualization of reward distributions over candidate trajectories, colored by reward value. 
Unsafe trajectories (e.g., collisions or off-road driving, red circles) are consistently assigned low rewards, while feasible trajectories near the ground truth (yellow circles) receive higher rewards, demonstrating that the reward-guided policy effectively suppresses unsafe behaviors while preserving multimodal diversity.}
\label{v3}
\end{figure}

Table~\ref{a_ft} evaluates different fine-tuning objectives. Compared to the model without fine-tuning, incorporating KTO with auxiliary supervision consistently improves PDMS. In particular, KTO complements the supervision loss by encouraging the policy to favor more desirable trajectories over undesirable ones, while the auxiliary supervision provides dense training signals for stable optimization. Moreover, reward supervision achieves the best performance (89.6), outperforming score supervision (89.0) used in Stage I. This suggests that the aggregated reward signal offers more task-aligned guidance, as it better matches the final trajectory selection criterion used during inference.

To better understand how the model ranks trajectory candidates, Fig.~\ref{v3} presents the reward distributions over predicted trajectories across diverse scenarios. The learned reward exhibits a clear separation between safe and unsafe trajectories while preserving multimodal feasible solutions. For instance, in the left-turn scenario (see Fig.~\ref{v3}(c)), trajectories corresponding to the intended turning behavior receive higher rewards, whereas alternative non-turning trajectories are consistently suppressed. Similar patterns are observed in Figs.~\ref{v3}(a, b, d), where trajectories leading to collision or off-road driving are assigned low rewards, while those closer to the ground truth obtain relatively higher rewards.

\section{Conclusion}
We present EMPlan, an efficient multi-modal trajectory planning framework based on a hybrid architecture and reward-guided fine-tuning. Our method enables real-time multi-modal trajectory generation while improving planning performance beyond pure imitation learning. Experiments on the non-reactive NAVSIM benchmark demonstrate that EMPlan achieves state-of-the-art performance with competitive inference efficiency.
A limitation, however, is that our reward-guided training relies on simulator-based evaluation, which prevents us from directly scaling to benchmarks that lack compatible simulation environments. 
 In future work, we plan to explore simulator-free reward modeling strategies to improve training scalability,  and we will also explore reactive closed-loop evaluation to better assess how the model behaves when other agents respond to its actions.



\bibliographystyle{IEEEtran}  
\bibliography{ref}     

\end{document}